\documentclass[sigconf,preprint]{acmart}
\renewcommand\footnotetextcopyrightpermission[1]{} 
\usepackage[utf8x]{inputenc}
\usepackage{xcolor}
\usepackage{xspace}
\usepackage{listings}
\usepackage{graphicx}

\usepackage{amsmath} \usepackage{amssymb} \usepackage{amsfonts} 
\usepackage{listings}
\usepackage{todonotes}
\usepackage{pgfplots}
\usepgfplotslibrary{groupplots}
\usetikzlibrary{pgfplots.groupplots}
\usetikzlibrary{plotmarks}

\newcommand{\tooltxt}{HyVarRec\xspace}
\newcommand{\tool}{{\sf \tooltxt}\xspace}

\makeatletter
\lst@Key{countblanklines}{true}[t]%
    {\lstKV@SetIf{#1}\lst@ifcountblanklines}

\lst@AddToHook{OnEmptyLine}{%
    \lst@ifnumberblanklines\else%
       \lst@ifcountblanklines\else%
         \advance\c@lstnumber-\@ne\relax%
       \fi%
    \fi}
\makeatother

\begin{document}

\title{Anomaly detection in Context-aware Feature Models}

\author{Jacopo Mauro}
\email{mauro@imada.sdu.dk}
\affiliation{  \institution{University of Southern Denmark}
}

\begin{abstract}

Feature Models are a mechanism to organize the configuration space
and facilitate the construction of software variants by describing
configuration options using features, i.e., a name representing a 
functionality. 
The development of 
Feature Models is an error prone activity and detecting their anomalies is a 
challenging and important task needed to promote their usage.

Recently, Feature Models have been extended with context to capture the 
correlation of configuration options 
with contextual influences and user customizations.
Unfortunately, this extension makes the task of detecting anomalies harder.
In this paper, we formalize the anomaly 
analysis in Context-aware 
Feature Models and we show how Quantified Boolean 
Formula (QBF) solvers can be used to detect anomalies without relying on 
iterative calls to a SAT solver. By extending the 
reconfigurator engine \tooltxt, we present findings evidencing 
that QBF solvers can outperform the common techniques for anomaly 
analysis.
\end{abstract}



\pagestyle{plain} 
\maketitle

\section{Introduction}
\label{section:intro}

Software Product Lines (SPLs) are a technology for large-scale reuse 
for a set of closely related software systems~\cite{Pohl:2005}, which allows 
companies to customize their software systems through configuration.
At the core of SPL engineering is the modeling of common and variable parts 
of software systems.
On the conceptual side, common and variable parts are described in 
terms of \emph{features}, which represent the configurable functionality of a 
system~\cite{Kang:1990}.
Features are often not independent on each other and to represent the relation 
between features a \emph{variability model} can be 
employed.
Among the most popular variability models are Feature 
Models (FMs)~\cite{Kang:1990} which often are represented visually in 
feature diagrams, a tree-like notation that structures features hierarchically.
 
\begin{figure}
\includegraphics[width=\linewidth]{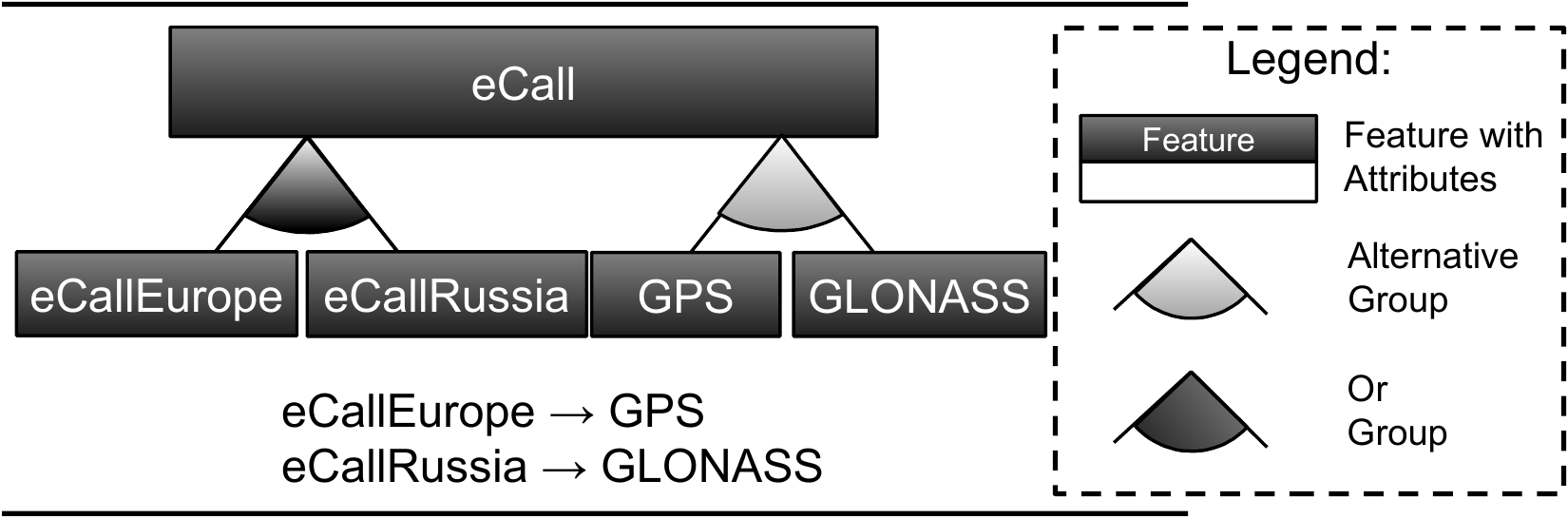}
\caption{Example of Feature Diagram.}
\label{fig:FM_example}
\end{figure} 
 
As an example, Figure \ref{fig:FM_example} shows the feature diagram that 
represents the emergency call feature \texttt{eCall} for a 
car~\cite{DBLP:conf/vamos/MauroNSY16}. The 
\texttt{eCall} service 
is provided by the \texttt{eCallEurope} or the \texttt{eCallRussia} feature. 
The \texttt{eCallEurope} relies on the positioning 
data of a \texttt{GPS} system while the \texttt{eCallRussia} service instead 
relies on 
positioning data 
provided by the \texttt{GLONASS} satellite system. The FM in 
Figure~\ref{fig:FM_example} graphically represents the dependencies between the 
features. In particular, to select the feature \texttt{eCall} the feature 
\texttt{eCallEurope} or the feature \texttt{eCallRussia} must be selected 
(or-group). 
Instead, it is required to select one and only one among the features 
\texttt{GPS} and \texttt{GLONASS} (alternative-group). Additional constraints, 
dubbed Cross-Tree Constraints (CTC),
are added. In 
Figure~\ref{fig:FM_example}, CTC are implications and 
require that when the feature \texttt{eCallEurope} (resp. \texttt{eCallRussia}) 
is selected, 
\texttt{GPS} (resp. \texttt{GLONASS}) is also selected.

A product (or configuration) of the FM is valid if it selects the 
features 
without contradicting any of the constraints 
imposed by the FM (both the structural tree constraints and the Cross-Tree 
Constraints). Hence, a valid product describes one member of 
the SPL on a 
conceptual level without regard to its implementation.
On the implementation side, features are realized using \emph{realization 
artifacts}, such as code or documentation artifacts.

Recently, in \cite{DBLP:journals/scp/MauroNSY18} the notion of FM has been 
extended to encompass 
the possibility to link the validity of one configuration to external 
factors. 
The new Context-aware Feature Models (CaFM) allows expressing SPL that 
are adaptable to the environment where they are deployed and take user 
preferences into 
account.
The idea behind CaFM is the possibility to use 
context variables to represent the external factors and impose 
constraints between the value of these 
variables and the features. For instance, for FM 
in Figure 
\ref{fig:FM_example}, let us assume that the car 
manufacturer by law has to provide 
in the cars sold in Russia and only in those the \texttt{eCallRussia} feature. 
To do so, the FM can 
be enriched by a new context variable \texttt{Location} that can be 
externally set to \texttt{True} if and only if the car is sold in Russia. 
To 
bound the context variable to the features, it is possible to add the 
constraint that imposes the selection of \texttt{eCallRussia} if and 
only if the \texttt{Location} is set to \texttt{True}.

In common SPL engineering, 
errors in the creation of (Ca)FMs may happen and for 
this reason, procedures to determine and explain them are essential. For 
example, it is of paramount importance to understand if a change of a (Ca)FM 
makes it void, i.e., does not allow the possibility to have a valid 
configuration.
Many anomaly analysis and tools have been proposed for FM 
\cite{Batory:2005:SPLC,explaining_anomalies,review_analysis_FM} but, as 
initially investigated in \cite{splc_short_paper}, the introduction of context 
brings some changes.
In particular, the voidness analysis that checks if a valid configuration is 
always possible becomes more complex.

Motivated by the lack of a formal treatment for the anomaly analysis of CaFM 
and its complexity, in this paper we first formally define the 
anomaly analysis for CaFM and their complexity. We then show how 
Quantified Boolean Formula (QBF) solvers can be used to detect anomalies 
without relying on (possibly exponential) invocations of SAT solvers. 
Preliminary results 
obtained by extending the \tool reconfigurator 
engine~\cite{DBLP:journals/scp/MauroNSY18} show that the usage 
of 
QBF solvers can improve over the naive strategy of using a SAT-based solver on 
randomly generated CaFMs.

\emph{Structure of the paper} In Section \ref{sec:background} we introduce FM 
and their anomaly analysis. In Section \ref{sec:cafm} we formalize the notion 
of 
CaFM while in Section \ref{sec:analysis} we formalize and 
describe how to perform their anomaly analysis. In Section 
\ref{sec:tool} we introduce the \tool 
reconfiguration engine and we present some initial findings proving that the 
usage of QBF can be useful for performing the anomaly analysis. We draw 
some concluding remarks in Section \ref{sec:conclusions}.

\section{Background}
\label{sec:background}

In this section, we recap the notion of Feature Model (FM). We 
then list 
the most common FM analysis and mention the solving technologies used to solve 
them.

Among the different representations of 
feature models presented 
in the literature (see, e.g., \cite{Batory:2005:SPLC}), in the 
following we adopt the \emph{propositional formula} representation 
that is not as visual as the feature diagram 
representation depicted in Figure \ref{fig:FM_example}, but allows a more 
concise formal treatment.

\begin{definition}
A Feature Model is a pair $\mathcal{M} = (\mathcal{F},\phi)$
where:
\begin{itemize}
\item[-]
$\mathcal{F}$ is a set of features, and
\item[-]
$\phi$ is a
propositional formula where the variables 
$x$ are feature names.
\[
\;\phi \; ::=\; x \;\vert\; \phi \wedge \phi \;\vert\; \phi \vee \phi \;\vert\; 
\phi \rightarrow \phi \;\vert\; \neg \phi.
\]
\end{itemize}
The propositional formula $\phi$ over a set of features $\mathcal{F}$, 
represents the feature models whose products are sets
$\{f_1,...,f_n\}\subseteq\mathcal{F}$ ($n\ge 0$) such that $\phi$ is
satisfied by assigning value true to the variables $f_i$
($1\le i\le n$) and false to all other variables.

\end{definition}

\begin{example}
\label{example:fm}
Consider the example introduced in Figure \ref{fig:FM_example}.
This FM is represented by the pair $(\mathcal{F},\phi)$ 
where 
\begin{align*}
\mathcal{F} =& \{ \texttt{eCall}, \texttt{eCallEurope}, \texttt{GPS}, 
\texttt{eCallRussia}, \texttt{GLONASS} \}\\
\phi =& \texttt{eCall} \wedge \\
& 
\texttt{eCall} \rightarrow (\texttt{eCallEurope} 
\vee \texttt{eCallRussia}) \wedge\\
& 
\texttt{eCall} \rightarrow (\texttt{\texttt{GPS}} 
\vee \texttt{GLONASS}) \wedge \neg (\texttt{\texttt{GPS}} 
\wedge \texttt{GLONASS}) \wedge\\
& \texttt{eCallEurope} \rightarrow \texttt{eCall} \wedge
\texttt{eCallRussia} \rightarrow \texttt{eCall} \wedge \\
&\texttt{GPS} \rightarrow \texttt{eCall} \wedge
\texttt{GLONASS} \rightarrow \texttt{eCall} \wedge \\
& \texttt{eCallEurope} \rightarrow 
\texttt{GPS} \wedge \texttt{eCallRussia} \rightarrow \texttt{GLONASS} \\
\end{align*}
The formula $\phi$ in the first line makes sure that the feature \texttt{eCall} 
is selected since it is the root of the FM. The second and third lines 
capture the constraints imposed by the or/alternative-groups.
The fourth and fifth lines state that \texttt{eCall} is the parent 
feature of 
\texttt{eCallEurope}, \texttt{eCallRussia}, \texttt{GPS}, \texttt{GLONASS} and, 
therefore, it must be selected 
if its children are selected. The last line reports the  
Cross-Tree Constraints.
\end{example}

Given a formal definition of FMs, we can introduce their anomaly 
analysis. In 
particular, for conciseness, we report only the major 
ones referring to the interested reader the survey 
\cite{review_analysis_FM} for more details.
\begin{itemize}
\item \emph{Valid Product}. Given a FM $(\mathcal{F},\phi)$ and a product 
$\mathcal{P} \subseteq\mathcal{F}$, it is checked if the FM is valid, 
i.e., if the literals in $\mathcal{P} \cup 
\{\neg f . f \in \mathcal{F} \setminus  \mathcal{P} \}$ satisfy $\phi$.
\item \emph{Voidness}. Given a FM $(\mathcal{F},\phi)$, it is checked if 
it
does not allow products, i.e., if $\phi$ is not satisfiable.
This is 
probably the most important analysis because having a void FM means that no 
possible implementation can be obtained for the SPL.

\item \emph{Dead Features}. Given a FM $(\mathcal{F},\phi)$ and a feature $f 
\in \mathcal{F}$, the feature $f$ is dead if it can never be 
selected, i.e., if there is no valid product 
that contains it or, alternatively, if $f \wedge \phi$ is not satisfiable. The 
dead 
feature analysis retrieves all the dead features of a FM.
Note that, in 
general, dead features should be avoided for maintainability purposes since 
they can never be used in a product.
\item \emph{False Optional Features}. In FM features can be marked as 
mandatory or optional.
\footnote{We would like to remark that the definition of 
optional feature varies in the literature. For instance in 
\cite{Batory:2005:SPLC} a feature is optional if it can be selected and 
deselected, while in FeatureIDE \cite{DBLP:journals/scp/ThumKBMSL14} a feature 
is optional if it can be deselected when its parent in the feature diagram 
representation is selected. In this paper, we abstract from these details by 
requiring the users to decide which features are defined as optional or not.} 
Given a FM 
$(\mathcal{F},\phi)$ and a feature $f 
\in \mathcal{F}$ marked as optional, the feature $f$ is false 
optional when it is available in every possible product, i.e., when $\neg f 
\wedge \phi$ is not satisfiable. Given a subset of features 
marked as optional, the false optional feature analysis retrieves all the 
false optional ones. In general, false optional should be marked as mandatory 
for maintainability purposes.
\item \emph{Redundancies.} Redundancies are constraints that do not 
add information over existing ones. Redundancies can decrease maintainability 
but can also improve 
the readability and comprehensibility of the model. Using the propositional 
formula representation, a FM has redundancies if the formula $\phi$ contains 
redundant clauses (i.e., clauses that can be removed without altering the set 
of all the products).

\end{itemize}
\noindent
From the computational complexity point of view, the analysis of checking the 
validity 
of a product is polynomial on the size of the formula $\phi$, while the 
other analysis are NP-hard or coNP-hard \cite{Papadimitriou}. 

Different 
tools are used to perform the analysis and among the complete ones (i.e., tools 
that can prove the existence or the non 
existence of an anomaly) often the most used ones involve propositional 
logic-based 
tools such as SAT solvers, binary decision diagram, 
Constraint 
Programming solvers, SMT solvers, or description logic reasoners 
\cite{review_analysis_FM}.

The most used tools are based on SAT solvers~\cite{sat_solvers}, i.e., tools 
that check whether a Boolean propositional formula is satisfiable returning an 
assignment that makes the formula true, if any. 
For FM analysis such as the voidness one, the SAT solver backend is called only 
once, while other analysis require more than one invocation. For example, 
to determine all the dead features, it is possible to iteratively call a 
backend 
solver for every feature of the FM.
\begin{lstlisting}[float,numbers=left,countblanklines=false,label=
listing:death,caption=Dead Features Analysis using a SAT solver.]
def dead_feature(~$\mathcal{F}$~,~$\phi$~)
  push(~$\phi$~)
  fs = ~$\mathcal{F}$~
  dead_features = ~$\emptyset$~
  while fs ~$\neq$~ ~$\emptyset$~:
    f = fs.pop
    push(f)
    if not checkSat():
      dead_features.add(f)
    else:
      fs = fs - getModel()
    pop()
  return dead_features
\end{lstlisting}
This task is facilitated in modern incremental SAT solvers that support the 
possibility to perform push 
and pop operations to dynamically stack and retract formulas, thus avoiding 
repeating already performed computations. To 
exemplify the procedure, consider the pseudocode of Listing 
\ref{listing:death} that implements the analysis of dead features. We assume 
that the SAT solver can offer the following primitives:
\begin{itemize}
 \item \lstinline{pop} and \lstinline{push} to remove and add a 
formula on the stack;
 \item \lstinline{checkSat} that checks if the 
conjunction of the formulas on the stack is satisfiable;
 \item \lstinline{getModel} that retrieves the set of the positive literals 
of the last satisfiable solution computed.
\end{itemize}

Given a FM $(\mathcal{F},\phi)$, the idea of this algorithm is to start with 
$\phi$ (Line 2) and then in a loop check if a given feature is dead. At 
every interaction of the loop, a feature is selected (Line 6), pushed to 
the stack (Line 7), and the SAT solver invoked to check if $\phi \wedge f$ is 
satisfiable (Line 7). If the formula is 
not satisfiable the feature added is dead and can be added to the set of dead 
features (Line 9). If not, at Line 11 the features 
that 
have been selected are removed from the set of the features to check (if 
a feature is selected is not dead). Finally, in Line 12, the 
formula added in Line 7 is removed from the stack so the procedure can continue 
checking the next feature, if any. Note that when structural information on the 
FM is known, it is possible to perform additional 
optimizations. For example, as done by 
\cite{DBLP:journals/scp/ThumKBMSL14}, when the diagram of the FM is known 
it is possible to mark as dead all the child features of a dead feature.

In the remaining part of the paper, we will use not only SAT solvers but also 
Quantified Boolean Formula 
(QBF) and Satisfiability Modulo Theory (SMT) solvers. QBS solvers \cite{qbflib} 
or QSAT solvers, generalize a SAT solver enabling to check the 
satisfiability of quantified Boolean formulas, thus 
being able to process formulas with universal quantifiers. SMT solvers 
\cite{smt} also extend SAT solvers by generalizing variables using  
predicates from 
a variety of underlying theories, thus allowing for instance to 
support integer variables and arithmetic constraints. These solvers are more 
powerful than SAT solvers and can be used to detect anomalies in just one 
invocation. In this paper, in particular, the experiments rely on the 
state-of-the-art 
SMT solver Z3\cite{z3}. This solver supports a 
huge variety of theories and can handle also formulas involving 
universal quantifiers, thus making it also a QBF solver. In particular, Z3 
uses several approaches to handle quantifiers 
like pattern/model-based 
quantifier instantiation or quantifier elimination
\cite{DBLP:conf/fmcad/WintersteigerHM10}.

\section{Context-aware Feature Models}
\label{sec:cafm}

In this section, we formalize the notion of Context-aware Feature Model (CaFM). 
Following \cite{DBLP:journals/scp/MauroNSY18}, a context can be 
considered as a variable that someone externally (e.g., the user of the 
software or the environment) can set. These new variables impose constraints 
over features and therefore on the products that can be obtained.

Without loss of generality,\footnote{In \cite{DBLP:journals/scp/MauroNSY18} 
context are variable that take values on finite domain. This can be easily 
modeled with a set of variables having a Boolean domain.} for presentation' 
sake, we can restrict ourselves to consider context variables that can take 
only 
two values: \texttt{true} or \texttt{false}. In this way, the notion of  
context and feature almost coincide, with the only difference that the value of 
features can be controlled by the developer of the CaFM, while the value of 
contexts is decided externally.

\begin{definition}
A Context-aware Feature Model (CaFM) is a tuple $(\mathcal{C}, 
\mathcal{F},\phi)$
where:
\begin{itemize}
\item[-]
$\mathcal{C}$ is a set of context,
\item[-]
$\mathcal{F}$ is a set of features, and
\item[-]
$\phi$ is a
propositional formula where the variables 
$x$ are feature or context names.
\[
\;\phi \; ::=\; x \;\vert\; \phi \wedge \phi \;\vert\; \phi \vee \phi \;\vert\; 
\phi \rightarrow \phi \;\vert\; \neg \phi.
\]
\end{itemize}
Fixed the value of the context variables in $\mathcal{C}$, a 
propositional formula $\phi$ over a set of features $\mathcal{F}$, 
represents the feature models whose products are sets
$\{f_1,...,f_n\}\subseteq\mathcal{F}$ ($n\ge 0$) such that $\phi$ is
satisfied by assigning value true to the variables $f_i$
($1\le i\le n$) and false to all other variables.
\end{definition}

\begin{example}
Consider the FM introduced in Figure \ref{fig:FM_example} and its 
propositional representation shown in Example \ref{example:fm}.
As discussed in Section \ref{section:intro}, imagine that the car manufacturer 
by law has to provide
in the cars sold in Russia the \texttt{eCallRussia} 
feature, otherwise the \texttt{eCallEurope} 
feature. To 
capture this situation, the FM can 
be enriched by a new context variable \texttt{Location} that externally can be 
set to \texttt{true} if the car is sold in Russia. Then the CaFM modeling this 
situation is the tuple $(\mathcal{C}, \mathcal{F},\phi')$ where 
\begin{align*}
\mathcal{C} =& \{ \texttt{Location} \}\\
\mathcal{F} =& \{ \texttt{eCall}, \texttt{eCallEurope}, \texttt{GPS}, 
\texttt{eCallRussia}, \texttt{GLONASS} \}\\
\phi' =& \phi \wedge \texttt{Location} \rightarrow \texttt{eCallRussia} 
\wedge\\
& \neg \texttt{Location} \rightarrow \neg \texttt{eCallRussia}
\end{align*}
where $\phi$ in the formula introduced in Example \ref{example:fm}.

The introduction of the context forbids the selection of 
\texttt{eCall\-Russia} 
when the context variable is set to \texttt{false}, while it forces its 
selection 
if \texttt{Location} is set to \texttt{true}. Therefore, the set of 
valid products available depends on the value of the context \texttt{Location}.

\end{example}

\section{Analysis of Context-aware Feature Models}
\label{sec:analysis}

In this section, we describe how the common analysis for FMs are lifted 
to the CaFM, discussing also how QBF 
solvers 
can be used to solve them when needed.

\subsection{Valid Product}

With the introduction of contexts, the natural extension of the product 
validity check is to verify if the 
product is valid in a given context.

\begin{definition}[Valid Product] Given a CaFM $(\mathcal{C}, 
\mathcal{F},\phi)$, a context assignment $\mathcal{D} \subseteq\mathcal{C}$ and 
a product 
$\mathcal{P} \subseteq\mathcal{F}$, $\mathcal{P}$ is valid in $\mathcal{D}$ 
when the literals in 
$ \mathcal{D} \cup \{\neg c . c \in \mathcal{C} \setminus  \mathcal{D} 
\} \cup \mathcal{P}  \cup
\{\neg f . f \in \mathcal{F} \setminus  \mathcal{P} \}$
satisfy $\phi$.
\end{definition}

It is easy to see that the introduction of context does not have 
a huge impact on the techniques used to decide whether a product is 
valid or not. Indeed, similarly to what happens for normal FMs, to validate a 
product of a CaFM the context must be fixed and, thus, 
the validation of the product requires checking 
if the ground formula of the CaFM is true. The complexity of this operation 
is therefore polynomial w.r.t. the size of the formula representing the CaFM.

\subsection{Voidness}

The 
straightforward way to extend the notion of voidness to the context-aware case 
is to say that a CaFM is void if there exists a context that does not 
admit a valid product.

\begin{definition}[Voidness]
A CaFM $(\mathcal{C}, 
\mathcal{F},\phi)$ is void  
if 
$\exists \mathcal{C} . \forall 
\mathcal{F} . \neg \phi$.
\end{definition}

Notation: given a 
set of variables $\mathcal{X} = \{x_i, \dots, x_n\}$ 
and a formula $\phi$, we write with $\exists \mathcal{X} . \phi$ the 
existential 
closure of $\phi$, i.e., $\exists x_1 . \dots \exists x_n . \phi$. 
Similarly, 
we shorten the formula $\forall x_1 . \dots \forall x_n . \phi$ with $\forall 
\mathcal{X} . \phi$.

From the definition of voidness, we can derive that checking it is a problem 
that 
belongs to the complexity class $\Sigma^P_2 = 
NP^{NP}$~\cite{Papadimitriou}, i.e., the 
class of problems that can be 
solved by calling a non-determinis\-tic polynomial time algorithm able to use a 
non-deterministic polynomial time oracle. 
Therefore, since checking the voidness of a FM is an NP-complete problem,
unless the polynomial hierarchy collapses, checking the voidness of a CaFM 
is more difficult than checking it for FM. As a 
consequence, a SAT solver is not enough to 
perform the 
voidness check in only one call, leaving the possibility of using a QBF solver 
instead. The implementation of the voidness check with a QBF solver is indeed
straightforward: 
QBF solvers support universally quantified formula and therefore it is possible 
to check if $\exists \mathcal{C} . \forall 
\mathcal{F} . \neg \phi$ is satisfiable directly.

The voidness of a CaFM can instead be checked by using iteratively a SAT 
solver as shown in Listing \ref{listing:fm_voidm}. The idea behind this 
procedure is to loop over all the possible context combinations and check if 
there exists one in which the resulting FM is void.
\begin{lstlisting}[float,numbers=left,countblanklines=false,label=
listing:fm_voidm,caption=FM Void Analysis using SAT solver]
def check(cs):
  if cs = []:
    if not checkSat():
      print("CaFM is void")
      exit(1)
  else:
    push(cs[0]) # context variable set to true
    check(cs[1:])
    pop()
    push(~$\neg$~cs[0]) # context variable set to false
    check(cs[1:])
    pop()

push(~$\phi$~)
check(~$\mathcal{C}$~)
print("CaFM not void")
\end{lstlisting}
In Listing~\ref{listing:fm_voidm} this is started by 
pushing 
on the stack the original formula $\phi$ (Line 13) and call 
the \lstinline{check} procedure (Line 14). The 
\lstinline{check} 
recursive procedure 
takes each context variable $c$ at the time and tries to set the context first  
to 
true (Line 7) and later to false (Line 10) to allow the exploration of all 
the possible 
combinations of 
contexts. The push and pop operation of 
the SAT 
solver are used to be able to reuse as much as possible the work already 
performed by the SAT solver. When all the context variables are grounded, 
the 
\lstinline{check} procedure checks if the conjunction of the formulas on the 
stack is satisfiable 
(Line 3). If not, a void FM has been found and the analysis can be 
stopped. If no void FM is found, the CaFM is not void and 
this can be returned (Line 15).

Note that the code in Listing \ref{listing:fm_voidm} performs 
up-to 
$2^{|\mathcal{C}|}$ invocations to the \lstinline{checkSat} procedure, which 
solves an 
NP-complete problem.

\subsection{Dead Features}

In CaFM, a straightforward extension of the dead feature concept is to consider 
a feature dead if it can not be selected in all the 
possible contexts. This leads to the 
following definition of dead feature.

\begin{definition}[Dead feature]
Given a CaFM $(\mathcal{C}, 
\mathcal{F},\phi)$, a future $f \in \mathcal{F}$ is dead if 
$\neg \exists \mathcal{C} \ . 
\ 
\exists \mathcal{F} \ . \ \phi \wedge f$.
\end{definition}

Similarly to what happens for the normal FM, this formula can be easily checked 
with a SAT solver.
To check all the features, the push and pop of the feature variables can be 
used to iteratively call a SAT solver as done in Listing \ref{listing:death}.
Hence, the problem of finding if a 
feature is dead is still a coNP-complete problem and checking if there 
is at least a dead feature is 
a problem that belongs to the class $\Delta_2^P = P^{NP}$, i.e., the 
class of problems that can be 
solved by calling a polynomial amount of times a non-deterministic polynomial 
time oracle like a SAT solver.

We would like to note that in case a QBF solver is available, it is possible 
to exploit it to find out if there is a dead feature in a (Ca)FM in just one 
call. This 
requires the addition of an auxiliary variable for every feature. 
Let us suppose that given a feature $f$ its fresh auxiliary 
variable is $\texttt{aux}(f)$ and 
$\mathcal{F}_{aux} = \{\texttt{aux}(f) . f \in \mathcal{F}\}$. A 
feature for a CaFM is dead if the following formula is satisfiable.

\[
\exists \mathcal{F}_{aux} \ . \ \texttt{OnlyOne}(\mathcal{F}_{aux}) \wedge
\forall \mathcal{C} \ . \
  \forall{F} \ . \ (\bigwedge_{f \in \mathcal{F}} \texttt{aux}(f) \rightarrow 
f) 
\rightarrow \neg \phi
\]
where
\[
\texttt{OnlyOne}(x_1, \dots, x_n) = 
\bigvee_{1 \leq i \leq n}  x_i \wedge
(\bigwedge_{1 \leq i,j \leq n, i\neq j} \neg x_i \vee \neg x_j)
\]

In this formula, intuitively, the 
auxiliary variable $\texttt{aux}(f)$ is used 
to force one feature $f$ to be always selected in the universally quantified 
formula. The \texttt{OnlyOne} predicate enforces one and only one 
auxiliary variable to be true. The universally 
quantified 
formula instead checks that for all the possible context and all the possible 
selections of features, the selection of the feature $f$ for which 
$\texttt{aux}(f)$ is true 
leads always to a void product. If this formula is satisfiable, then at 
least one $\texttt{aux}(f)$ is true and the corresponding feature $f$ is 
a dead feature.

Differently from the standard analysis exemplified in Listing 
\ref{listing:death}, another possible
strategy to locate dead features could be to let the SAT solver 
guide the search and prune the features that are not dead as shown in Listing 
\ref{listing:pruning}.
\begin{lstlisting}[float,numbers=left,countblanklines=false,label=
listing:pruning,caption=Dead Features Anlysis guided by a SAT 
solver.]
push(~$\phi$~)
fs = ~$\mathcal{F}$~
while fs ~$\neq$~ ~$\emptyset$~:
  push(~$\bigvee$~ fs)
  if not checkSat():
    return fs # fs are all dead features
  fs = fs ~$\setminus$~ getModel()   
return ~$\emptyset$~
\end{lstlisting}
The idea behind this algorithm is to start with the original formula $\phi$ and 
add the disjunction of features that have not been proven yet 
to belong to a product (Line 4). When the call to the SAT solver (Line 3) 
proves that the formula is unsatisfiable, none of the 
features added in the last iteration of the while loop can be selected and 
therefore all of them are returned 
as dead feature (Line 6). Otherwise, the selected feature can be excluded 
(Line 7) and 
the process can continue until either we reach an unsatisfiable formula or we 
prune all the features to check.

\subsection{False Optional Features}

For CaFM, the notion of false optional is naturally extended as a 
feature that i) is marked as optional and ii) for all the context and all the 
products 
it can not be deselected. Formally:

\begin{definition}[False optional]
Given a CaFM $(\mathcal{C}, 
\mathcal{F},\phi)$, a future $f \in \mathcal{F}$ marked as optional is 
false optional if $\neg \exists \mathcal{C} \ . \ 
\exists \mathcal{F} \ . \ \phi \wedge \neg f$.
\end{definition}

Similarly to what is done for detecting dead features, checking all the false 
optional feature can be done by calling iteratively a SAT solver, either by 
removing one feature at the time or by using the pruning technique presented in 
Listing \ref{listing:pruning}. The complexity of finding false features in CaFM 
is the same as the one for FM.

Even in this case, it is also possible to use a quantifier solver to detect if 
there are any false optional features in just one call.
Let us assume that that given a feature $f$ marked as optional  
$\texttt{aux}(f)$ is an auxiliary variable. Let us consider 
$\mathcal{F}_{aux}$ the set containing all the auxiliary variables 
corresponding to features marked as optional. The existence of a false optional 
feature can be proven by checking the satisfiability of the following formula.

\[
\exists \mathcal{F}_{aux} \ . \ \texttt{OnlyOne}(\mathcal{F}_{aux}) \wedge
\forall \mathcal{C} \ . \
  \forall{F} \ . \ (\bigwedge_{f \in \mathcal{F}} \texttt{aux}(f) \rightarrow 
\neg f) 
\rightarrow \neg \phi
\]

\subsection{Redundancies}

A constraint is redundant if it can be  
removed without altering the set of valid products for all the possible 
context combinations.
Formally:
\begin{definition}[Redundancy]
Given a CaFM $(\mathcal{C}, 
\mathcal{F},\phi \wedge \phi')$, $\phi$ is redundant 
if $\exists \mathcal{C} . \exists \mathcal{F} . \neg(\phi' \rightarrow \phi)$ 
is 
unsatisfiable.
\end{definition}

As a consequence, testing whether an instance does not 
contain any 
redundant clause is an NP-complete problem 
\cite{DBLP:conf/ecai/Liberatore02}\footnote{
Note that modern SAT solvers often remove automatically some for of 
redundancy as a pre-processing step \cite{DBLP:conf/cpaior/FourdrinoyGMS07}. 
They can therefore detect automatically some redundancies without 
degradation of performances.} and the same techniques used to check 
redundancies for FMs can be 
adopted for CaFMs.

\section{Evaluation}
\label{sec:tool}

In this section, we briefly introduce the CaFM analyzer \tool and present 
the findings on the usage of a QBF solver for the analyses of 
randomly 
generated CaFM.

\subsection{\tooltxt}

\tool \cite{DBLP:journals/scp/MauroNSY18} is a reconfiguration engine for 
CaFM that relies on the SMT solver Z3 \cite{z3} to find a valid product of a 
CaFM that minimizes (or maximizes) 
user-defined metrics. Originally intended for the reconfiguration of CaFM 
in presence of context changes, \tool has been extended to support the 
voidness check \cite{splc_short_paper} and later 
used 
by DarwingSPL\footnote{\url{https://gitlab.com/DarwinSPL/DarwinSPL}} to 
provide explanations for anomalies~\cite{DBLP:conf/gpce/NiekeMSTYF18}.

We have extended \tool to support the checking of the dead and false optional 
features by using Z3's QBF solver and the SAT guided pruning technique 
introduced in Listing \ref{listing:pruning}.
In the following we use the term \emph{Iterative} to refer to 
approaches that call iteratively a SAT solver (e.g., Listing 
\ref{listing:death} and \ref{listing:fm_voidm}), \emph{Forall} for the 
approaches that uses the QBF solver, and \emph{Pruning} for the 
approach that uses a SAT solver to guide the pruning of the features for the 
feature analysis check (e.g., Listing \ref{listing:pruning}).
For the \emph{Forall} approach, \tool uses the default tactics implemented by 
Z3 to solve 
quantified formulas.\footnote{\url{
https://stackoverflow.com/questions/20682763/z3-does-qe-tactic-preserve-equivale
nce-or-only-equisatisfiability/20719090}}
The feature analysis is performed by first trying to find the dead features and 
later the false optional ones. As an optimization, the features that 
during the detection of dead features were found to be deselected for a valid 
product were not 
considered for the false feature detection. The dead and false positive feature 
analysis can be 
stopped by \tool as soon as one anomaly is discovered or when all anomalies 
are 
detected.

\tool is publicly available at \url{https://github.com/HyVar/hyvar-rec} and 
supports the possibility to define features encoded 
either by integer values or 
directly as 
Boolean values. \tool also supports the possibility to use attributes and 
context 
variables that can take values in finite 
domain integer sets.

\subsection{Methodology}

To the best of our knowledge, there is no
established benchmark for CaFMs.
To be able to compare the different approaches of anomaly explanation we have 
created a CaFM 
random instance 
generator\footnote{\url{
https://github.com/HyVar/hyvar-rec/tree/master/test/cafm_generator}} by relying 
on a SAT formula generator 
\cite{DBLP:conf/aaai/0001KRS17}\footnote{The SAT generator is available online 
at \url{https://github.com/RalfRothenberger/Power-Law-Random-SAT-Generator}.}.
Given a target number of context variables and features, the SAT 
formula generator was used to generate propositional formulas with clauses 
having 3 literals taking the variables according to the uniform distribution. 
According to the target number of context, some literals were considered 
context variables while the others were considered features. The number of 
clauses of the propositional formula (and therefore its hardness) was 
controlled by a parameter expressing the ratio between the number of clauses 
and the number of features. The clauses having only context variable literals, 
if any, were removed to avoid restricting the possible space of context 
combinations.

We run a first set of experiments considering CaFM having 250 features 
and varying the number 
of context variables in the set $\{ 6, 8, 10, 12, 14 \}$. For every 
number in this set, we 
generated 100 instances, varying the ratio of clauses to number of features 
between 5 and 6. For the feature anomaly analysis, we stopped as soon 
as 
an anomaly was detected and we marked as optional all the features of the CaFM. 
We repeated every experiment 10 times.
All the experiments were run in a Docker container\footnote{The Docker image 
used for the tests is 
available at 
\url{https://hub.docker.com/repository/docker/jacopomauro/hyvar-rec}.} on a 
virtual machine having 2 vCPU and 8 GB of RAM. The Docker process was 
terminated if the entire execution was taking more than 6 GB of RAM or running 
for more than 5 minutes.

A second set of experiments were run by fixing the context number to 10 and 
varying the number of features in the set $\{200, 300, 350 \}$ following 
the same execution modalities and limitation of the first set of experiments.

\subsection{Results}

\begin{table}
 \begin{tabular}{|c|c|c|c|c|}
 \hline
Contexts & Result & Forall & Iterative & Total \\
\hline
\hline
6 & Void & 11 & 34 & 45 \\
 & Not Void & 12 & 43 & 55 \\
\hline
8 & Void & 25 & 31 & 56  \\
 & Not Void & 1 & 43 & 44  \\
\hline
10 & Void & 26 & 20 & 46  \\
 & Not Void & 2 & 52 & 54  \\
\hline
12 & Void & 31 & 31 & 62  \\
 & Not Void & 1 & 37 & 38  \\
\hline
14 & Void & 33 & 25 & 58  \\
 & Not Void &  & 42 & 42  \\
\hline
Total &  & 142 & 358 & 500  \\
\hline
 \end{tabular} 
\caption{Number of times a voidness check approach 
is the best by varying the number of context 
variables (features number = 250).}
\label{tab:nodominance_void}
\end{table}

\begin{table}
 \begin{tabular}{|c|c|c|c|c|c|}
 \hline
Contexts & Result & Forall & Iterative & Pruning & Total \\
\hline
\hline
6 & Dead & 5 & 29 &  & 34\\
 & False & 7 & 21 & 6 & 34\\
 & No Anomaly & 9 & 17 & 6 & 32 \\
\hline
8 & Dead & 4 & 26 &  & 30\\
 & False & 10 & 15 & 4 & 29\\
 & No Anomaly & 13 & 22 & 6 & 41 \\
\hline
10 & Dead & 9 & 19 &  & 28\\
 & False & 11 & 10 & 2 & 23\\
 & No Anomaly & 20 & 23 & 6 & 49 \\
\hline
12 & Dead & 4 & 22 &  & 26\\
 & False & 11 & 8 & 2 & 21\\
 & No Anomaly & 17 & 28 & 8 & 53\\
\hline
14 & Dead & 6 & 18 &  & 24\\
 & False & 6 & 4 & 5 & 15\\
 & No Anomaly & 16 & 33 & 12 & 61 \\
\hline
Total &  & 148 & 295 & 57 & 500 \\
\hline
 \end{tabular} 
\caption{Number of times a feature anomaly approach is the best by 
varying the number 
of 
context (features number = 250).}
\label{tab:nodominance_feature}
\end{table}

The most interesting finding of our experiments is that no single approach 
dominates the other. Table \ref{tab:nodominance_void} reports for instance the 
number of times an approach is the best (i.e., lower average running time) 
for the voidness 
analyses in the first set of experiments. It is possible to see that the 
Iterative approach is often the best, 
being the fastest in 358 cases. However, in 142 instances the Forall 
approach 
performs 
better (especially if the instance is 
void). This can be partially explained by a conjecture of the developers of 
Z3 that stated that ``using quantifiers is a good 
option only if a very 
small percentage of the instances are needed to show that a problem is 
unsatisfiable''.\footnote{
https://stackoverflow.com/questions/13268394/avoiding-quantifiers-in-z3} It 
may be the case that the heuristics of the QBF solver led to the finding of 
the void context early on, compared to the iterative search approach in which 
the combinations of context are tried in a specific order.

As far as the feature analysis is concerned, Table \ref{tab:nodominance_feature}
shows the number of times every approach was the best. Also in this case, the 
Iterative approach is usually better being the fastest 
in 295 instances, but there is a non-negligible number of cases in which the 
Forall and Pruning approach are the best. The Forall approach seems to be 
competitive with the 
Iterative one for instances that have no dead features but false optional ones.
Pruning is usually slower than the other two approaches and is not competitive 
in case there are dead features. We conjecture that this is due to the fact 
that the Pruning approach requires to identify all the dead features to 
return one.

In the second set of experiments, as 
expected, we noticed that the 
more feature there are the higher are the average running times. The 
approaches start to timeout for some instances already when 300 features are 
considered. In 
particular, for the feature check the Iterative approach timeouts for 12 
instances (i.e., at least one in 10 repetitions took more than 300 seconds), 
the Forall timeouts for 1 instance, and the Pruning for 2. For the voidness 
check, the Forall timeouts in 5 instances with 300 features while the Iterative 
timeouts for the first time only for instances with 350 features.

\begin{figure}

\begin{tikzpicture}
  \colorlet{vbs}{red}
  \colorlet{iterative}{black}
\begin{groupplot}[ 
        group style={
        group size=1 by 5,
        vertical sep=35pt,
        },
    width=.5\textwidth,
]
\nextgroupplot[
  ylabel=Time (s),
  ylabel near ticks,
  width=.5\textwidth,
  height=4cm,
  title=\textbf{6 Context},
]
\addplot[draw=vbs] table[x expr=\coordindex+1, 
y=VBS6]{plots/250_features_validate.dat}; 
\addplot[draw=iterative, densely dotted] table[x expr=\coordindex+1, 
y=Iterative6]{plots/250_features_validate.dat}; 
\nextgroupplot[
  ylabel=Time (s),
  ylabel near ticks,
  width=.5\textwidth,
  height=4cm,
  title=\textbf{8 Context},
]
\addplot[draw=vbs] table[x expr=\coordindex+1, 
y=VBS8]{plots/250_features_validate.dat}; 
\addplot[draw=iterative, densely dotted] table[x expr=\coordindex+1, 
y=Iterative8]{plots/250_features_validate.dat}; 
\nextgroupplot[
  ylabel=Time (s),
  ylabel near ticks,
  width=.5\textwidth,
  height=4cm,
  title=\textbf{10 Context},
]
\addplot[draw=vbs] table[x expr=\coordindex+1, 
y=VBS10]{plots/250_features_validate.dat}; 
\addplot[draw=iterative, densely dotted] table[x expr=\coordindex+1, 
y=Iterative10]{plots/250_features_validate.dat}; 
\nextgroupplot[
  ylabel=Time (s),
  ylabel near ticks,
  width=.5\textwidth,
  height=4cm,
  title=\textbf{12 Context},
]
\addplot[draw=vbs] table[x expr=\coordindex+1, 
y=VBS12]{plots/250_features_validate.dat}; 
\addplot[draw=iterative, densely dotted] table[x expr=\coordindex+1, 
y=Iterative12]{plots/250_features_validate.dat}; 
\nextgroupplot[
  ylabel=Time (s),
  ylabel near ticks,
  xlabel=Instances,
  width=.5\textwidth,
  height=4cm,
  title=\textbf{14 Context},
  legend style = {
    column sep = 10pt,
    legend columns = 1,
    legend to name = grouplegend,}]

]
\addplot[draw=vbs] table[x expr=\coordindex+1, 
y=VBS14]{plots/250_features_validate.dat};
\addlegendentry{best of iterative \& forall}
\addplot[draw=iterative, densely dotted] table[x expr=\coordindex+1, 
y=Iterative14]{plots/250_features_validate.dat}; 
\addlegendentry{iterative}
\end{groupplot}

\node at (group c1r5.south) [anchor= north, yshift=-1cm] 
{\ref{grouplegend}};

\end{tikzpicture}
\caption{Solving times for voidness analysis with 250 features and varying 
number of 
context variables.}
\label{fig:250_voidness}
\end{figure}
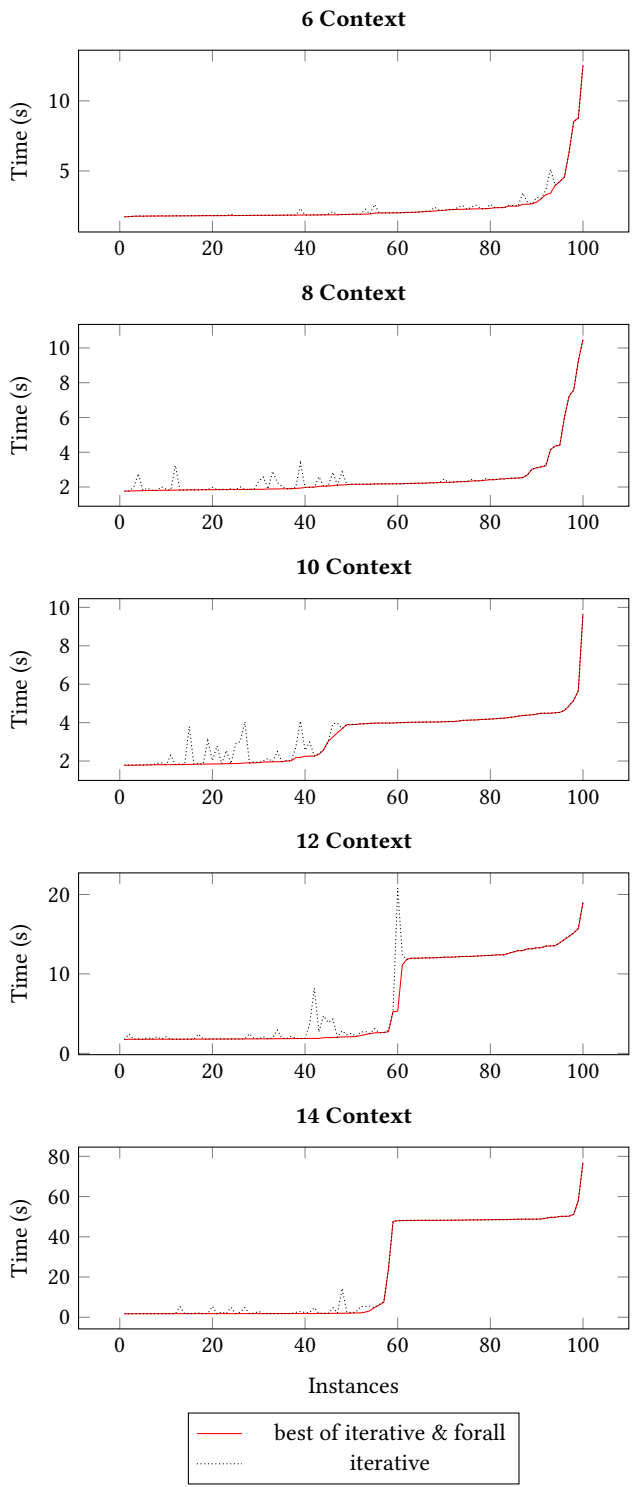

\begin{figure}

\begin{tikzpicture}
  \colorlet{vbs}{red}
  \colorlet{iterative}{black}
  \colorlet{forall}{blue}
\begin{groupplot}[ 
        group style={
        group size=1 by 5,
        vertical sep=35pt,
        },
    width=.5\textwidth,
]
\nextgroupplot[
  ylabel=Time (s),
  ylabel near ticks,
  width=.5\textwidth,
  height=4cm,
  title=\textbf{6 Context},
]
\addplot[draw=vbs] table[x expr=\coordindex+1, 
y=VBS6]{plots/250_features_features.dat}; 
\addplot[draw=forall,densely dashed] table[x expr=\coordindex+1, 
y=WithForall6]{plots/250_features_features.dat}; 
\addplot[draw=iterative, densely dotted] table[x expr=\coordindex+1, 
y=Iterative6]{plots/250_features_features.dat};
\nextgroupplot[
  ylabel=Time (s),
  ylabel near ticks,
  width=.5\textwidth,
  height=4cm,
  title=\textbf{8 Context},
]
\addplot[draw=vbs] table[x expr=\coordindex+1, 
y=VBS8]{plots/250_features_features.dat}; 
\addplot[draw=forall,densely dashed] table[x expr=\coordindex+1, 
y=WithForall8]{plots/250_features_features.dat}; 
\addplot[draw=iterative, densely dotted] table[x expr=\coordindex+1, 
y=Iterative8]{plots/250_features_features.dat};
\nextgroupplot[
  ylabel=Time (s),
  ylabel near ticks,
  width=.5\textwidth,
  height=4cm,
  title=\textbf{10 Context},
]
\addplot[draw=vbs] table[x expr=\coordindex+1, 
y=VBS10]{plots/250_features_features.dat}; 
\addplot[draw=forall,densely dashed] table[x expr=\coordindex+1, 
y=WithForall10]{plots/250_features_features.dat}; 
\addplot[draw=iterative, densely dotted] table[x expr=\coordindex+1, 
y=Iterative10]{plots/250_features_features.dat};
\nextgroupplot[
  ylabel=Time (s),
  ylabel near ticks,
  width=.5\textwidth,
  height=4cm,
  title=\textbf{12 Context},
]
\addplot[draw=vbs] table[x expr=\coordindex+1, 
y=VBS12]{plots/250_features_features.dat}; 
\addplot[draw=forall,densely dashed] table[x expr=\coordindex+1, 
y=WithForall12]{plots/250_features_features.dat}; 
\addplot[draw=iterative, densely dotted] table[x expr=\coordindex+1, 
y=Iterative12]{plots/250_features_features.dat}; 
\nextgroupplot[
  ylabel=Time (s),
  ylabel near ticks,
  xlabel=Instances,
  width=.5\textwidth,
  height=4cm,
  title=\textbf{14 Context},
  legend style = {
    column sep = 10pt,
    legend columns = 1,
    legend to name = grouplegend2,}]

]
\addplot[draw=vbs] table[x expr=\coordindex+1, 
y=VBS14]{plots/250_features_features.dat}; 
\addlegendentry{best of iterative \& forall \& pruning}
\addplot[draw=forall,densely dashed] table[x expr=\coordindex+1, 
y=WithForall14]{plots/250_features_features.dat}; 
\addlegendentry{best of iterative \& forall}
\addplot[draw=iterative, densely dotted] table[x expr=\coordindex+1, 
y=Iterative14]{plots/250_features_features.dat}; 
\addlegendentry{iterative}
\end{groupplot}

\node at (group c1r5.south) [anchor= north, yshift=-1cm] 
{\ref{grouplegend2}};

\end{tikzpicture}
\caption{Solving time for feature anomaly analysis with 250 features and 
varying number of context variables.}
\label{fig:250_features}
\end{figure}
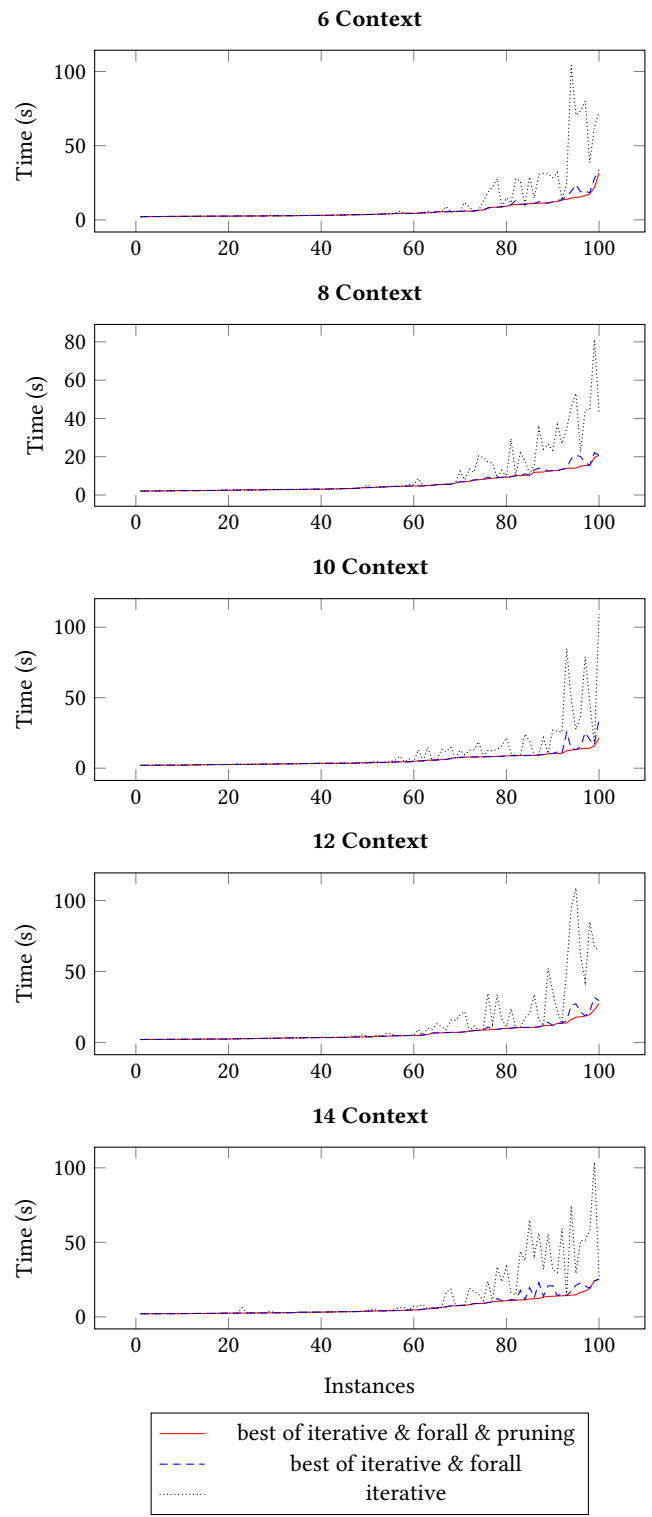

To better visualize the improvement in solving times when multiple 
approaches can be used, we have plotted in Figure~\ref{fig:250_voidness} the 
average times taken by the Iterative approach (black dotted line) and the times 
taken by the fastest among the Iterative and the Forall approaches (continuous 
red line) for the voidness analysis.
The plots are related to the first set of experiments using instances having 
250 features and varying the number of contexts. For every context, the 
instances have been sorted by the average solving time. 
Figure \ref{fig:250_features} present instead the similar plots 
depicting the average solving time for the feature anomaly analysis, 
considering the Iterative approach (black 
dotted line), the best among the Iterative approach and 
the Forall approaches (blue dashed line), and also the best among all the three 
approaches used (continuous red line).

As far as the voidness analysis is concerned, when more than 10 
context are used it is possible to see that there is a jump in the average 
solving times. This is mainly due to the fact that detecting an anomaly for 
void instances is often faster than proving that there are no anomalies. The
majority of the instances before the jump in solving times are therefore void 
instances. Note, however, that there are exceptions. For example, considering 
the instances having 12 contexts, the most difficult 2 instances for the 
Iterative approach are void (and luckily in one of these two cases the 
Forall approach is better, reducing the solving time from 20 to 5 seconds as 
shown by the spike in the 12 context plot of Figure 
\ref{fig:250_voidness}).
Overall, it is possible to see that the Forall approach can improve the solving 
times, especially when the instance is void.

The improvement of performances is even more visible for the feature anomaly 
analysis. As can be seen from Figure \ref{fig:250_features}, there are 
occurrences where the Iterative 
approach takes more than 100 seconds while the 
Forall or Pruning take less than half time. While for the voidness 
analysis the performance gains are usually on the easier void instances,
here the gains are also on the more difficult instances.

\begin{figure}

\begin{tikzpicture}
  \colorlet{vbs}{red}
  \colorlet{iterative}{black}
  \colorlet{forall}{blue}
\begin{groupplot}[ 
        group style={
        group size=1 by 2,
        vertical sep=35pt,
        },
    width=.5\textwidth,
]
\nextgroupplot[
  ylabel=Time (s),
  ylabel near ticks,
  width=.5\textwidth,
  height=4cm,
  title=\textbf{200 features},
]
\addplot[draw=vbs] table[x expr=\coordindex+1, 
y=VBSv]{plots/200f_10c_void.dat}; 
\addplot[draw=iterative, densely dotted] table[x expr=\coordindex+1, 
y=gridv]{plots/200f_10c_void.dat};
\nextgroupplot[
  ylabel=Time (s),
  ylabel near ticks,
  xlabel=Instances,
  width=.5\textwidth,
  height=4cm,
  title=\textbf{350 Features},
  legend style = {
    column sep = 10pt,
    legend columns = 1,
    legend to name = grouplegend4,}]

]
\addplot[draw=vbs] table[x expr=\coordindex+1, 
y=VBSv]{plots/350f_10c_void.dat}; 
\addlegendentry{best of iterative \& forall \& pruning}
\addplot[draw=iterative, densely dotted] table[x expr=\coordindex+1, 
y=gridv]{plots/350f_10c_void.dat}; 
\addlegendentry{iterative}

\end{groupplot}

\node at (group c1r2.south) [anchor= north, yshift=-1cm] 
{\ref{grouplegend4}};

\end{tikzpicture}
\caption{Solving times for voidness analysis with 10 contexts and varying 
number of 
features.}
\label{fig:200350void}
\end{figure}
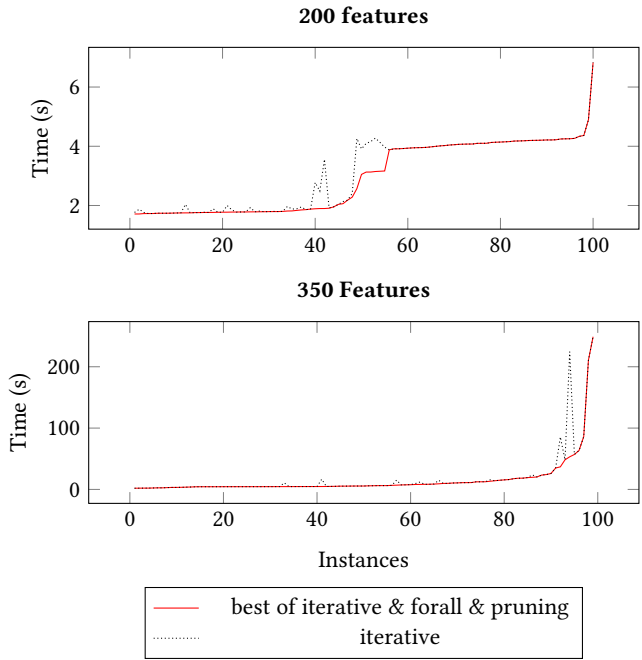

\begin{figure}

\begin{tikzpicture}
  \colorlet{vbs}{red}
  \colorlet{iterative}{black}
  \colorlet{forall}{blue}
\begin{groupplot}[ 
        group style={
        group size=1 by 2,
        vertical sep=35pt,
        },
    width=.5\textwidth,
]
\nextgroupplot[
  ylabel=Time (s),
  ylabel near ticks,
  width=.5\textwidth,
  height=4cm,
  title=\textbf{200 features},
]
\addplot[draw=vbs] table[x expr=\coordindex+1, 
y=VBSf]{plots/200f_10c_features.dat}; 
\addplot[draw=forall,densely dashed] table[x expr=\coordindex+1, 
y=withForallf]{plots/200f_10c_features.dat}; 
\addplot[draw=iterative, densely dotted] table[x expr=\coordindex+1, 
y=gridf]{plots/200f_10c_features.dat};
\nextgroupplot[
  ylabel=Time (s),
  ylabel near ticks,
  xlabel=Instances,
  width=.5\textwidth,
  height=4cm,
  title=\textbf{350 features},
  legend style = {
    column sep = 10pt,
    legend columns = 1,
    legend to name = grouplegend5,}]

]
\addplot[draw=vbs] table[x expr=\coordindex+1, 
y=VBSf]{plots/350f_10c_features.dat}; 
\addlegendentry{best of iterative \& forall \& pruning}
\addplot[draw=forall,densely dashed] table[x expr=\coordindex+1, 
y=withForallf]{plots/350f_10c_features.dat}; 
\addlegendentry{best of iterative \& forall}
\addplot[draw=iterative, densely dotted] table[x expr=\coordindex+1, 
y=gridf]{plots/350f_10c_features.dat}; 
\addlegendentry{iterative}
\end{groupplot}

\node at (group c1r2.south) [anchor= north, yshift=-1cm] 
{\ref{grouplegend5}};

\end{tikzpicture}
\caption{Solving time for feature anomaly analysis with 10 contexts and 
varying number of features.}
\label{fig:200350feature}
\end{figure}
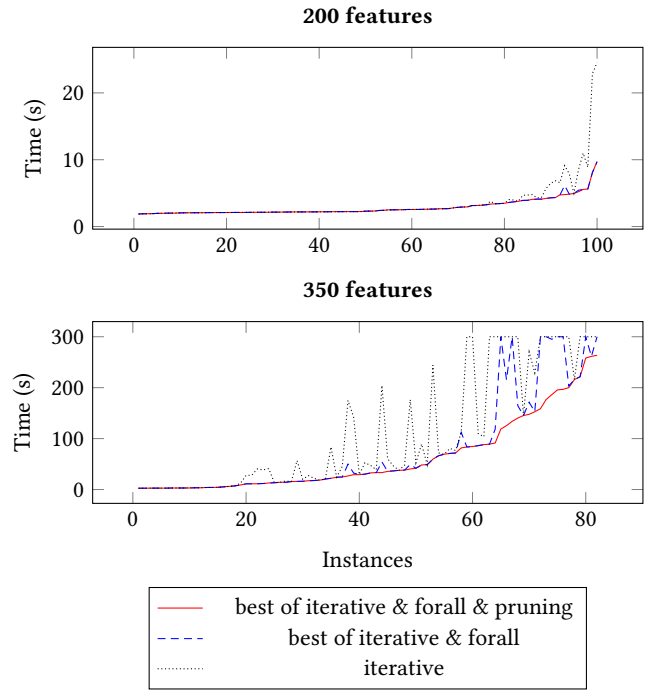

For the second set of experiments, Figures 
\ref{fig:200350void} and \ref{fig:200350feature} present the average solving 
times for the instances having 10 contexts and varying the number of features.
For space reason, we just present the plots for the experiments having 200 and 
350 features.
Since when using 350 features some instances timeout, for these plots we take 
into account only the instances that were solved by at least 
an approach within the timeout. In case of an approach timeouts for an 
instance, we consider its average time equal to the timeout (300 
seconds).

Looking at the plots in Figure \ref{fig:200350void}, it is possible to see 
that with 350 features the jump in the average 
solving time is less pronounced. We believe that this is due probably to the 
randomic nature of the instances: when more features are used there is a 
higher chance to generate a more difficult void 
instance w.r.t. to easy non void instances. For small instances the Forall 
approach seems to bring benefits for the detection of void 
instances that are, however, solved 
fast also by the Iterative approach. On the contrary, for bigger instances, 
the Forall approach can bring benefits also for more 
difficult non void instances (e.g., the big spike in the 350 feature  
plot in Figure \ref{fig:200350void} is due to a reduction of solving time 
of more than 100 seconds obtained by the Forall approach on a non void 
instance).

Similar results can be observed in Figure \ref{fig:200350feature} for the 
feature anomaly analysis. In this case, in particular, it is possible to see 
that for big instances the Forall and 
Pruning approach allow to solve instances that otherwise would have not been 
solved. There are many instances where the Forall or Pruning approach are able 
to solve the instance 100 seconds faster than the Iterative approach.

\begin{figure}
\begin{tikzpicture}
\begin{groupplot}[ 
        group style={
        group size=1 by 2,
        vertical sep=35pt,
        },
    width=.5\textwidth,
]
\nextgroupplot[
  ylabel=Time (s),
  ylabel near ticks,
  width=.5\textwidth,
  height=7cm,
  title=\textbf{Iterative approach},
]
\addplot[draw=blue,error bars/.cd, y dir=both,y explicit] table[x 
expr=\coordindex+1, 
y=gridv_avg, y error=std_gridv]{plots/variability_250_grid_void.dat}; 
\nextgroupplot[
  ylabel=Time (s),
  ylabel near ticks,
  xlabel=Instances,
  width=.5\textwidth,
  height=7cm,
  title=\textbf{Forall approach},
]
\addplot[draw=blue,error bars/.cd, y dir=both,y explicit] table[x 
expr=\coordindex+1, 
y=Forall, y error=std]{plots/variability_250_forall_void.dat}; 
\end{groupplot}
\end{tikzpicture}

\caption{Solving times with standard deviation for voidness analysis with 250 
features. Only instances taking in average more than 10 seconds 
and with all the 10 repetitions below 300 seconds are plotted.}
\label{fig:variability_void}
\end{figure}

\begin{figure}
\begin{tikzpicture}
\begin{groupplot}[ 
        group style={
        group size=1 by 3,
        vertical sep=35pt,
        },
    width=.5\textwidth,
]
\nextgroupplot[
  ylabel=Time (s),
  ylabel near ticks,
  width=.5\textwidth,
  height=7cm,
  title=\textbf{Iterative approach},
]
\addplot[draw=blue,error bars/.cd, y dir=both,y explicit] table[x 
expr=\coordindex+1, 
y=grida_avg, y error=std_grida]{plots/variability_250_grid_features.dat}; 
\nextgroupplot[
  ylabel=Time (s),
  ylabel near ticks,
  width=.5\textwidth,
  height=7cm,
  title=\textbf{Forall approach},
]
\addplot[draw=blue,error bars/.cd, y dir=both,y explicit] table[x 
expr=\coordindex+1, 
y=foralla_avg, y error=std_foralla]{plots/variability_250_forall_features.dat}; 
\nextgroupplot[
  ylabel=Time (s),
  ylabel near ticks,
  xlabel=Instances,
  width=.5\textwidth,
  height=7cm,
  title=\textbf{Pruning approach},
]
\addplot[draw=blue,error bars/.cd, y dir=both,y explicit] table[x 
expr=\coordindex+1, 
y=pruninga_avg, y 
error=std_pruninga]{plots/variability_250_pruning_features.dat}; 
\end{groupplot}
\end{tikzpicture}
\caption{Solving times with standard deviation for feature anomaly analysis 
with 250 
features. Only instances taking in average more than 10 seconds 
and with all the 10 repetitions below 300 seconds are plotted.}
\label{fig:variability_features}
\end{figure}

We would like to conclude this section by also addressing the variability of the 
various approaches. We noticed that the 
Forall approach in general and the Iterative approach for the feature analyses 
have a big variability (i.e., standard deviation sometimes superior to 
50\%).

Figure \ref{fig:variability_void} presents the average solving times and their 
standard deviation for the Iterative and Forall approaches for the voidness 
analysis. In these plots, for presentation sake, we considered all the 
instances used in the first set of experiments with 250 features and plotting 
only the ones with an average solving time of more than 10 seconds and in which 
all the 10 repetitions ended within the timeout. From the error bars that 
represent the standard deviation, it is easy to see how the Forall approach 
times vary while the Iterative approach has a lower variability (i.e., less than 
5 \%). We believe that this is due to the randomized nature of the QBF solver.

Figure \ref{fig:variability_features} presents the average solving 
times and their 
standard deviation for the Iterative, Forall, and Pruning approaches for the 
feature anomaly analysis. As before, we considered all the 
instances used in the first set of experiments with 250 features and plotting 
only the ones with an average solving time of more than 10 seconds and in which 
all the 10 repetitions ended within the timeout. Also in this case the Forall 
approach shows a significant performance variability. 
Surprisingly, also the Iterative approach manifests a 
high variability. We believe that this is due to the fact that the Iterative 
approach prunes features 
depending on the solutions found 
by the SMT solver. The solutions found vary based on some random 
internal choices, thus probably causing the variability of the times of the 
Iterative approach. The Pruning approach instead has a lower variability. This 
is probably due to the fact that to find a single anomaly the Pruning 
approach has to account for either all the dead features or all the false 
optional features.

Note that the number of instances considered in the plot for the Pruning 
approach in Figure~\ref{fig:variability_features} is considerably higher than 
the ones for the other two approaches. This is due to the fact that in average 
the Pruning approach is slower than the Iterative and the Forall ones, and 
therefore it took more than 10 seconds for more instances.

\section{Related Works and Conclusions}
\label{sec:conclusions}

To the best of our knowledge, this is the first paper to formally define 
CaFMs and their anomaly analysis. CaFMs were originally introduced in 
\cite{DBLP:journals/scp/MauroNSY18} and the idea that QBF solvers can be 
used to check the voidness was first presented in 
\cite{splc_short_paper} without providing, however, any comparison w.r.t. the 
iterative approach.
In \cite{DBLP:conf/gpce/NiekeMSTYF18} \tool has been extended to 
provide the explanation of voidness and perform the feature analysis, but 
only the iterative approach was used. We are not aware of 
other tools that perform analysis of context-aware SPLs. The closest 
approaches to ours for the configuration is the UbiFex notation to model 
context-aware SPLs~\cite{Fernandes:2008} that allows them to determine using a 
simulator if the FM is void given a certain context. In particular, in 
that work their emphasize that it is hard to 
reason for each possible context regarding the FM being void.
Other works which focus on modeling context-aware SPLs exist.
For instance, in \cite{context_MSPL} context-awareness is captured by 
providing a second FM while in 
\cite{context_ontology_spl} ontologies are used instead to model the context. 
Unfortunately, these works just present these models without discussing their 
analysis.

In this paper, we describe how anomaly analysis can be performed for 
CaFMs. In general, the analysis techniques relying on iterative calls to a 
SAT 
solver can still be used, but initial findings show that the 
usage of QBF 
solver in parallel to the standard techniques can improve the overall 
performance. We also formalize a strategy for performing future 
analysis by letting the SAT solver guide the pruning of the features. Despite 
this technique requires finding 
first all the feature anomalies before providing one, sometimes it performs 
even better than other approaches that stop as soon as the first anomaly has 
been detected.

We have implemented the new analysis approaches in \tool and we have evaluated 
them on randomly generated instances. Our goal was to try to compare these 
strategies on a uniform framework such as the one provided by the SMT solver 
Z3. 
Comparing the performance by using different SAT/QBF solvers along the lines 
of \cite{DBLP:conf/vamos/SpreySKNMTS20} is left as future work. We expect 
that the adoption of SAT solvers can improve the performance on certain 
instances, especially considering that often they are more effective in getting 
the model when a solution for a formula has been found. For 
example, when Z3 is used 
for large FM in FeatureIDE \cite{DBLP:journals/scp/ThumKBMSL14}
or \tool, the time it takes to retrieve the 
model of a satisfiable formula can outweighing the cost of pushing and popping 
every 
single feature and check only the satisfiability. Clearly, using SAT solvers 
directly can therefore potentially speed up the solving time.

The results presented in this paper strongly 
depend on the 
random 
instances considered in the experiments. It may be 
that the structure of the real world instances is different from the randomic 
instances we generate. For this reason we consider this results only 
preliminary and we hope that new industrial benchmarks for CaFMs 
will be created to 
validate the performance of the current strategies on real instances. 
Moreover, compared to advanced tools designed for having in input FM diagrams  
such as 
\cite{DBLP:journals/scp/ThumKBMSL14}, \tool can not perform advanced 
optimizations to prune features based on the feature diagram notation.
It is left as future work to see whether advance pruning techniques 
based on 
the FM structure can be adopted to speed up the search when the CaFM feature 
diagram is available.

Due to the fact that all the problem considered (except 
product validity) are NP-hard~\cite{np_hard} and that
there is no approach (e.g., SAT, SMT) that 
dominates the others for solving such problems, we believe that the results 
obtained are not surprising. Often the 
performance of one solver may vary also by orders of magnitude depending on the 
instance to be solved or the random seed used. This can be exploited by 
\emph{Algorithm Portfolios} 
\cite{DBLP:journals/ai/GomesS01,DBLP:conf/lopstr/AmadiniGM15} that, based on 
the instances to solve, run different approaches to obtain an overall better 
solver. As done for the SAT and Constraint Programming fields 
\cite{xu2008satzilla,malitsky2013algorithm,sunnycp,sunnycp2,isac,smac}, further 
studies are needed to be able to devise strategies to select promising 
approaches 
based on the instances to solve.

\begin{acks}
The author would like to thank Michael Nieke
for providing precious feedback on a preliminary version of this document.
\end{acks}




\end{document}